\pdfoutput=1
\documentclass[11pt]{article}

\usepackage[utf8]{inputenc}
\usepackage[T1]{fontenc}
\usepackage{lmodern}

\usepackage{amsmath,amssymb,amsfonts}
\usepackage{graphicx}      % figures
\usepackage{natbib}        % bibliography
\usepackage{booktabs}
\usepackage{geometry}
\usepackage{hyperref}

\title{Human Imitated Bipedal Locomotion with\\
Frequency Based Gait Generator Network}

\author{
  Yusuf Baran Ates \quad Omer Morgul\\[4pt]
  Dept. Electrical \& Electronics Eng., Bilkent University,\\
  Ankara 06800 T\"{u}rkiye\\
  \texttt{baran.ates@bilkent.edu.tr, morgul@ee.bilkent.edu.tr}
}

\date{}  % no date on title page

\begin{document}

\maketitle

\begin{abstract}
Learning human-like, robust bipedal walking remains difficult due to hybrid dynamics and terrain variability. We propose a lightweight framework that combines a gait generator network learned from human motion with Proximal Policy Optimization (PPO) controller for torque control. Despite being trained only on flat or mildly sloped ground, the learned policies generalize to steeper ramps and rough surfaces. Results suggest that pairing spectral motion priors with Deep Reinforcement Learning (DRL) offers a practical path toward natural and robust bipedal locomotion with modest training
cost.
\end{abstract}


%===============================================================================
\section{Introduction}

Bipedal locomotion remains one of the most challenging problems in robotics due to its hybrid dynamics, balance constraints, and the need for smooth transitions for real world deployment. Classical control approaches such as the Linear Inverted Pendulum (LIPM) and Spring-Loaded Inverted Pendulum (SLIP) models have provided numerical solutions to gait stability and energy exchange, \cite{slip}, but they often lack adaptability and robustness on irregular terrains. Another approach was to use state-based reference tracking with manually tuned phase variables as the SIMBICON framework, \cite{yin2007simbicon}. Optimization studies using Covariance Matrix Adaptation Evolution Strategy (CMA-ES) based extensions enabled walking at variable speeds and slopes by tuning reference trajectories, \cite{wang2009optimizing}. However these controllers are dependent on predefined gait states and environmental conditions, which result in fragile systems in unseen environments.

Bio-inspired approaches have also gained attention for generating rhythmic and adaptive walking behaviors. Central Pattern Generators (CPGs), \cite{taga1991self}, provide biologically plausible oscillatory control structures. \cite{ccatalbacs2022two} showed that recurrent neural networks trained on CPG-generated trajectories can outperform the original oscillators, producing smoother transitions and more stable gait control.

DRL has recently emerged as a powerful alternative, capable of synthesizing locomotion policies directly from interaction. RL-based frameworks have achieved robust walking and running on platforms such as Cassie and Digit,~\cite{li2025reinforcement,krishna2021learning,surveypaper}. However, many methods still rely on a single reference motion or a fixed gait library,~\cite{itahashi2024reinforcement,xie2018feedback}, restricting motion diversity and human-likeness. To overcome this, imitation-driven methods such as DeepMimic,~\cite{peng2018deepmimic} and Generative Motion Priors,~\cite{zhang2025natural} introduced MoCap or generative reference data to produce smoother and more natural trajectories. Multi-speed human imitation frameworks have further extended these results to various walking speeds using limited motion data,~\cite{su2023multi}. \cite{peng2025gait} proposed a curriculum learning strategy with predefined gait conditions, allowing a single recurrent policy to learn standing, walking, running, and smooth transitions without relying on motion capture data.

Motivated by these developments, this work investigates a frequency-based reinforcement learning framework for human-like bipedal walking. The proposed controller combines frequency-parameterized gait generation inspired by CPG dynamics with DRL to learn adaptive control policies. By integrating imitation weight decay and Random State Initialization (RSI) methods, gait conditioned rewards, and speed-conditioned objectives, the framework aims to achieve robust and natural locomotion across multiple speeds and terrains. The results demonstrate that combining frequency-driven imitation with reinforcement learning provides a simple and effective pathway toward natural and resilient bipedal gait generation.

%===============================================================================
\section{Bipedal Controller}

We introduce a \textbf{frequency-based gait generator network} trained on human motion-capture data transformed into the Fourier domain. 
This transformation encodes periodic gait patterns, allowing the network to generate smooth, speed-conditioned joint trajectories from minimal input (leg length and desired velocity). 
We then integrate gait generator network into a torque-controlled \textbf{PPO}, \cite{schulman2017proximal}, agent trained using General Advantage Estimation, \cite{gae}, that learns to track reference motions in simulation. 

Pybullet is used as simulation source. Experiments demonstrate that the proposed imitation-guided PPO achieves stable and human-like walking across speeds of $0$–$2.2\,\text{m/s}$, ramp angles from $-15^{\circ}$ to $+15^{\circ}$, and terrain perturbations up to $10\,\text{cm}$. Compared to pure DRL baselines, the approach converges faster, exhibits smoother joint coordination, and maintains robustness under noise and slope variations.

\subsection{Data Preparation}

We train gait generator network on \cite{schreiber2019multimodal}. The dataset covers five walking regimes with 1143 gait trials from 50 participants (24F/26M, $37.0\!\pm\!13.6$\,yrs, $1.74\!\pm\!0.09$\,m, $71.0\!\pm\!12.3$\,kg). Unlike many MoCap datasets that re-target all motions to a single canonical skeleton, this work preserves the original body dimensions of the users. This is crucial for speed-conditioned generalization: the mapping from speed to joint phase/frequency depends on morphology, so learning directly from individualized limb lengths produces references that better match commanded speeds on novel morphologies.

Markers on the hip/knee/ankle are mapped to six joint angles (bilateral hip, knee, ankle). Foot–strike events define gait cycles; each cycle is labeled with average Center of Mass (CoM) velocity. Raw 100\,Hz trajectories are low–pass filtered (5\,Hz Butterworth) and downsampled to 10\,Hz. Variable-length cycles are unified via a 32-point FFT (real + imaginary parts retained). This \emph{frequency–domain} encoding yields a fixed-size, periodic representation; inverse FFT reconstructs 3.2\,s cycles (32 frames), sufficient to capture the slowest speeds.

\subsection{Gait Generator Network Training}

Each trial provides inputs $x = [\,v,\ \ell_r,\ \ell_l\,]$ (desired speed and right/left leg lengths) and output $\hat{Q}\in\mathbb{R}^{32\times 6}$ (joint angles reconstructed from the 32-point FFT). The network learns
\begin{equation}
    f_\theta:\ x\ \mapsto\ \hat{Q}
\end{equation}
where $\theta$ represents the learnable parameters of the related neural network, see Fig. \ref{controlloop}. We perform a grid search over hidden layers $\{1,2,3\}$, widths $\{128,256,512\}$, learning rates $\{10^{-3},10^{-4},10^{-5}\}$, batch sizes $\{32,64,128\}$, and ReLU/Tanh activations. Each configuration is trained for up to 10{,}000 epochs with early stopping, and repeated with 5 random seed. The best model is found to be a 2-layer MLP (512 units, ReLU), Adam with $10^{-4}$ learning rate, 64 batch size. 

\subsection{Simulation Environment and Robot Model}
We use a 7 Degree of Freedom (DoF) planar biped (torso–hip joint plus bilateral hip, knee, ankle). We use the model  with the same parameters as given in \cite{yin2007simbicon}. A combined HAT (head–arms–trunk) model is used, following common practice. All experiments are done in the Pybullet simulation environment.

\subsection{Observation, Action, and Control Scheme}

\begin{figure} 
\centering 
\includegraphics[width=1.0\textwidth]{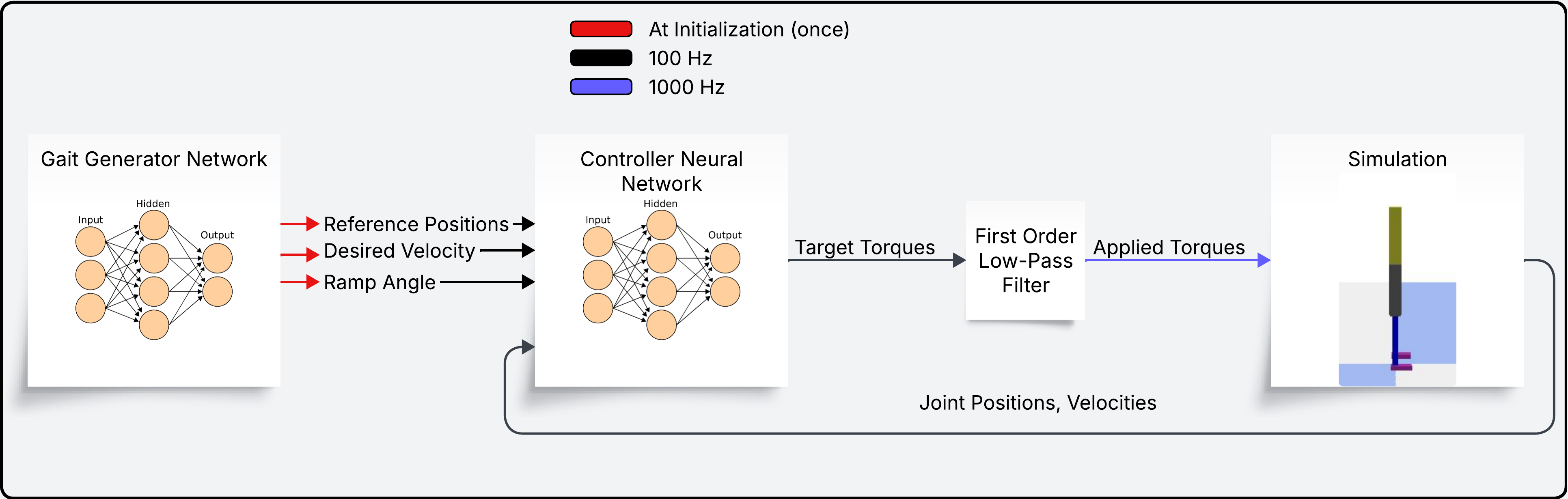}
\caption{Controller Loop.} 
\label{controlloop} 
\end{figure}

We empirically select a compact history+preview observation that improved training stability inspired by~\cite{peng2020learning}. The final observation is 55\,D:
\[
s_t=\bigl[v_{ds},\,r_{\text{angle}},\,v_{com},\,
q_{t-1},\,q_t,\,\hat{q}_t,\,\hat{q}_{t+1},\,\hat{q}_{t+10},\,
a_{t-1},\,\dot{q}_t,\,\hat{\dot{q}}_t\bigr],
\]
with component sizes are given in Table~\ref{tab:obs_components}. (Lower-body references are 6-D; actual vectors are 7-D.)

\begin{table}[t]
\centering
\caption{Observation $s_t$ components (total 55\,D)}
\label{tab:obs_components}
\begin{tabular}{lcc}
\toprule
Component & Dim. & Description \\
\midrule
$v_{ds}$ & 1 & commanded forward speed (scalar) \\
$r_{\text{angle}}$ & 1 & ramp pitch angle (rad) \\
$v_{com}$ & 1 & measured CoM forward speed \\
$q_{t-1},\,q_t$ & $7+7$ & past \& current joint angles \\
$\hat{q}_t,\,\hat{q}_{t+1},\,\hat{q}_{t+10}$ & $6+6+6$ & current/preview references \\
$a_{t-1}$ & 7 & previous torques \\
$\dot{q}_t,\,\hat{\dot{q}}_t$ & $7+6$ & actual \& reference joint vels \\
\bottomrule
\end{tabular}
\end{table}

Actions are desired torques for all 7 joints. The simulator runs at 1\,kHz; the high-level policy runs at 100\,Hz, and actions pass through a first-order low-pass filter to avoid jerky behavior (Fig.~\ref{controlloop}).

\subsection{Agent Training}
\label{sec:ppo_training}

We optimize a torque policy with PPO,~\cite{schulman2017proximal}. Training proceeds for 15\,M steps. The training domain randomizes \emph{ramp inclinations in $[-5^\circ, +5^\circ]$}, flat terrain, and target speeds $v^\star\!\in\![0,2.0]$\,m/s. Steeper slopes and noisy planes are held out for evaluation.

\begin{table}[t]
\centering
\caption{PPO hyperparameters}
\label{tab:ppo_params}
\begin{tabular}{lcc}
\toprule
Parameter & Value & Notes \\
\midrule
Learning rate & $3\!\times\!10^{-4}\!\to\!1\!\times\!10^{-4}$ & decays linearly \\
Entropy coef.\ $\alpha$ & $1\!\times\!10^{-3}\!\to\!1\!\times\!10^{-4}$ & decays linearly \\
Batch size & 256 & minibatch/epoch \\
Rollout horizon $n_{\text{steps}}$ & 8192 & on-policy window \\
PPO epochs $n_{\text{epochs}}$ & 5 & per update \\
Clip range $\varepsilon$ & 0.15 & trust region \\
Total timesteps & 15M & \\
\bottomrule
\end{tabular}
\end{table}

\begin{figure}[t] 
\centering 
\includegraphics[width=0.75\textwidth]{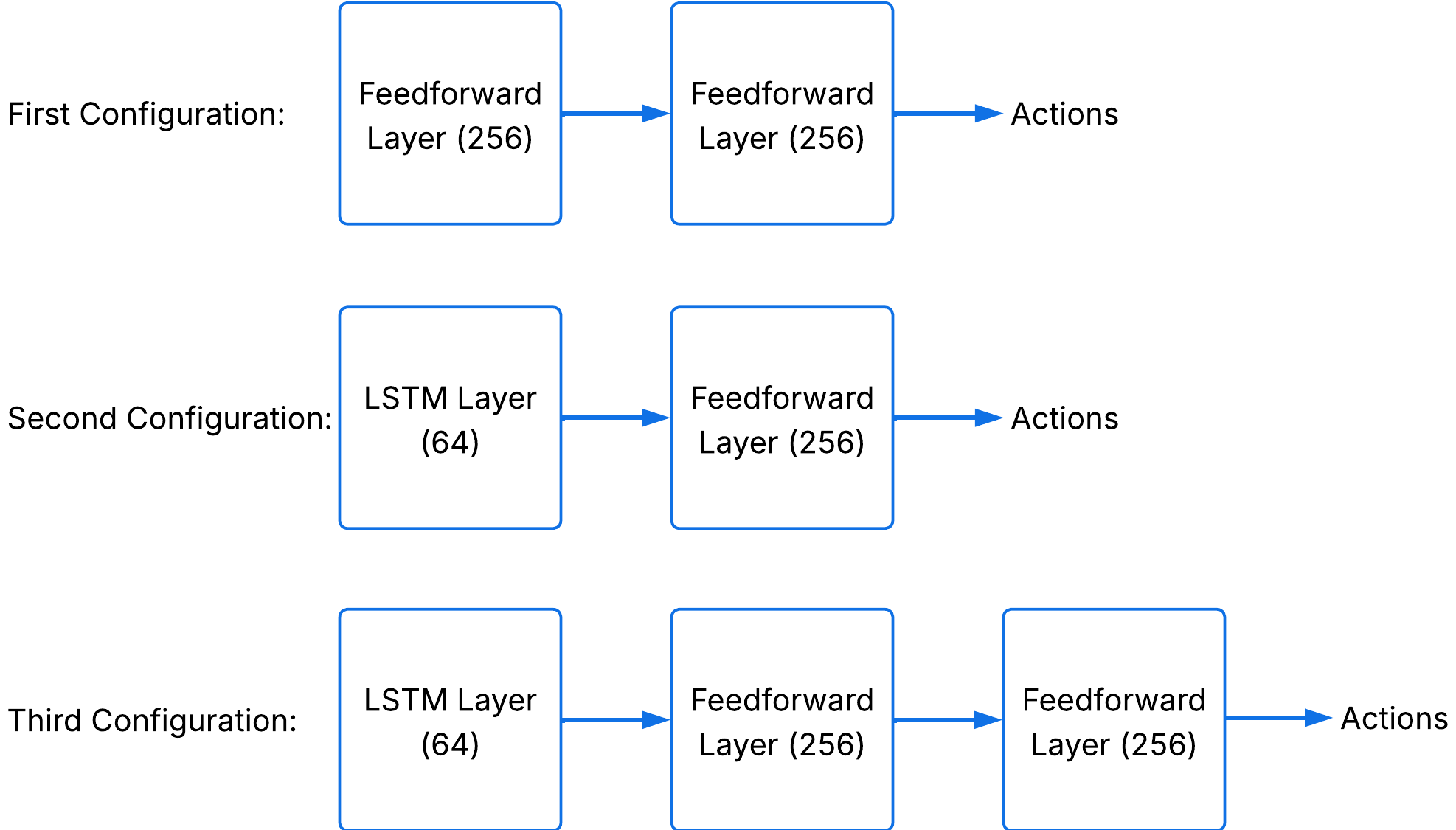} 
\caption{Policy Network Architectures} 
\label{configs}
\end{figure}

We train 3 different neural network configurations in identical settings that can be seen in Fig. \ref{configs}.

RSI method is used at the beginning of each episode, the pose of the robot is set to a randomly selected state from the reference motion. as in \cite{peng2018deepmimic}  to avoid excessive exploration of the poor states. Early termination is used and the reward is set to huge negative number if the robot's height decreases from some threshold or the angle of the torso exceeds some threshold. Imitation decay technique is also used to gradually relax the imitation reward so the policy can deviate when beneficial for stability/efficiency. The reward function is defined as:
\begin{equation}\label{reward_func}
    r(t)=\omega_{imitation}(t)\,r_{imitation}(t)+\omega_{gait}(t)\,r_{gait}(t)
\end{equation}
with schedules
\begin{equation}
\begin{split}
\omega_{\text{imitation}}(t) &= 1 - \alpha\!\left(\tfrac{\text{current step}}{\text{total step}}\right)\\
\omega_{\text{gait}}(t) &= 1 + \alpha\!\left(\tfrac{\text{current step}}{\text{total step}}\right)
\end{split}
\end{equation}
where $\alpha\!\in\![0,1)$ controls the trade-off and step is the total training step taken. This schedule preserves human-like structure early, then prioritizes task/robustness later. Without motion decay, the weights remain fixed as $\omega_{imitation}=1$ and $\omega_{gait} = 1$. 

\subsection{Reward Design}
The reward function (\ref{reward_func}) consists of 2 main terms. The fist term
encourages the model to follow the trajectories provided by the gait
generator network. The second term encourages the model to
follow the desired velocity while increasing the gait stability.

We match hips/knees strongly and ankles more softly, in both position and velocity. Ankles are highly contact adaptive that responds to local slope, surface noise, and impact, so over-constricting them harms the ground interaction in ramps and rough terrain. By contrast, hip/knee trajectories primarily govern CoM progression and stance height and are less sensitive to small contact perturbations. Let $q_t,\hat q_t,\dot q_t,\hat{\dot q}_t$ denote actual/reference joint positions/velocities. The imitation reward is constructed as follows:
\begin{equation}
\begin{aligned}
r_{\text{imitation}}(t) &= r_{\text{hip}}(t) + r_{\text{knee}}(t) + r_{\text{ankle}}(t), \\[6pt]
r_{\text{hip}}(t) &= 0.75\,e^{-5 \sqrt{\,\Delta q_{r\!h}^2 + \Delta q_{l\!h}^2\,}}
                  + 0.15\,e^{-0.2 \sqrt{\,\Delta \dot{q}_{r\!h}^2 + \Delta \dot{q}_{l\!h}^2\,}}, \\[6pt]
r_{\text{knee}}(t) &= 0.75\,e^{-5 \sqrt{\,\Delta q_{r\!k}^2 + \Delta q_{l\!k}^2\,}}
                  + 0.15\,e^{-0.2 \sqrt{\,\Delta \dot{q}_{r\!k}^2 + \Delta \dot{q}_{l\!k}^2\,}}, \\[6pt]
r_{\text{ankle}}(t) &= 0.25\,e^{-5 \sqrt{\,\Delta q_{r\!a}^2 + \Delta q_{l\!a}^2\,}}
                   + 0.05\,e^{-0.2 \sqrt{\,\Delta \dot{q}_{r\!a}^2 + \Delta \dot{q}_{l\!a}^2\,}}.
\end{aligned}
\end{equation}
where $\Delta q$/$\Delta\dot q$ denote the reference versus actual position and velocity errors for the indicated bilateral joints where the following subscripts $_l,_r,_h,_k,_a$ represent left, right, hip, knee and ankle, respectively. For the imitation reward parameter, the same intuition is followed as \cite{peng2020learning}. The gait reward is given as:
\begin{equation}\label{gait_rew}
    r_{\text{gait}}(t)=r_{\text{alive}}(t)+r_{\text{contact}}(t)+r_{\text{speed}}(t)-r_{\text{torque}}(t).
\end{equation}
The gait reward promotes safe posture, speed tracking, contact consistency, and energy use, where the individual reward parameters in (\ref{gait_rew}) are defined below. 

With CoM height $z_{com}$ and torso angle $q_{\text{torso}}$, $r_{alive}$ is defined as follows:
\begin{equation}
\small
\begin{aligned}
r_{\text{alive}} &=
\begin{cases}
+0.5, & \text{if } z_{\text{com}} \in [0.95,\, 1.25], \\[4pt]
-0.5, & \text{if } z_{\text{com}} \in (1.25,\, 1.45] \text{ or } [0.75,\, 0.95), \\[4pt]
-100, & \text{if } |q_{\text{torso}}| > 0.9 \text{ or } z_{\text{com}} > 1.25 \text{ or } z_{\text{com}} < 0.75.
\end{cases} \\[6pt]
\text{done} &=
\begin{cases}
\text{True}, & \text{if } |q_{\text{torso}}| > 0.9 \text{ or } z_{\text{com}} > 1.25 \text{ or } z_{\text{com}} < 0.75, \\[4pt]
\text{False}, & \text{otherwise.}
\end{cases}
\end{aligned}
\end{equation}
Episode is terminated early when done flag is raised. With finite-difference velocity $\dot x_t = (x_t - x_{t-\Delta})/\Delta t$ and $\Delta t=10$\,ms. $r_{speed}$ is defined as follows:
\begin{equation}
r_{\text{speed}}(t)=0.6\,e^{-3\,|\dot{x}_t-\dot{x}^\star|}.
\end{equation}

Feet are defined with four contact points. Four gait phases are defined for double support, right swing, left swing and flight. Gait phase $(\phi)$ is inferred from the vertical force each foot encounters from the ground.
\begin{equation}
\phi =
\begin{cases}
\text{double support}, & \text{if } F_R\geq 10\,N \text{ and } F_L \geq 10\,N, \\[3pt]
\text{right swing}, & \text{if } F_R < 10\,N \text{ and } F_L \geq 10\,N, \\[3pt]
\text{left swing}, & \text{if } F_R \geq 10\,N \text{ and } F_L < 10\,N. \\[3pt]
\text{flight}, & \text{if } F_R < 10\,N \text{ and } F_L < 10\,N.
\end{cases}
\end{equation}

$F_L,F_R,c_L,c_R,x_L,x_R$ is the vertical forces, active contact counts and the lateral position of the left and right foot. $z_{\text{plane}}(x)$ is the height of the plane at lateral position $x$. Desired swing clearance is $0.15$\,m to support high swing clarence. With $\sigma$ being the sigmoidal function, $r_{contact}$ is defined as follows:
\begin{equation}
\resizebox{\columnwidth}{!}{$
r_{\text{contact}}(t)=
\begin{cases}
\sigma\!\bigl(2[(c_L{+}c_R)-4]\bigr), &\phi= \text{double support},\\
\frac{1}{2}\sigma\!\bigl(-20|z_R-z_{\text{plane}}(x_R)-0.15|\bigr)+\frac{1}{2}\sigma\!\bigl(2(c_L-2)\bigr), &\phi= \text{right swing},\\
\frac{1}{2}\sigma\!\bigl(-20|z_L-z_{\text{plane}}(x_L)-0.15|\bigr)+\frac{1}{2}\sigma\!\bigl(2(c_R-2)\bigr), &\phi= \text{left swing},\\
0, & \text{flight}.
\end{cases}
$}
\end{equation}
To minimize the energy usage $r_{torque}$ is defined as follows:
\begin{equation}
    r_{\text{torque}}(t)=\tfrac{1}{1000}\sum_j^7 |a_j|.
\end{equation}
Where $a_j$ corresponds to torque applied at $j$'th joint. 

To test the capability of the model, we evaluate (i) \textbf{speed matching} (0–2.2 \,m/s), (ii) \textbf{ramp/rotation} ($-15^\circ$ to $+15^\circ$), and (iii) \textbf{noisy planes} (height perturbations up to 20 \,cm). (iv) \textbf{velocity tracking} (track the changing velocity in one episode) Ablation studies are performed on  \textbf{RSI} (on/off) and \textbf{imitation decay} (on/off).

%===============================================================================
\section{Results}

\begin{table*}[t]
\centering
\caption{Performance comparison across all configurations. 
Lowest MSE and highest success/range values are highlighted in bold.}
\resizebox{\textwidth}{!}{
\begin{tabular}{lcccccc}
\toprule
\textbf{Configuration} & 
\textbf{Velocity Difference MSE} $\downarrow$ & 
\textbf{Success (Rotation)} $\uparrow$ & 
\textbf{Success (NP0)} $\uparrow$ & 
\textbf{Success (NP1)} $\uparrow$ & 
\textbf{Range (NP0)} $\uparrow$ & 
\textbf{Range (NP1)} $\uparrow$ \\
\midrule
Configuration 1              & 0.030900 & 0.923 & \textbf{0.654} & \textbf{0.826} & \textbf{8.057} & 8.586 \\
Configuration 2              & 0.037228 & 0.776 & 0.416 & 0.501 & 7.779 & 8.204 \\
Configuration 3              & 0.052145 & 0.843 & 0.455 & 0.562 & 6.226 & 6.665 \\
Configuration 1 (Decay $\alpha{=}0.25$) & \textbf{0.003632} & 0.853 & 0.465 & 0.584 & 6.644 & 7.002 \\
Configuration 1 (Decay $\alpha{=}0.5$)  & 0.029028 & \textbf{0.962} & 0.471 & 0.598 & 6.297 & 6.847 \\
Configuration 1 (No RSI)     & 0.060353 & 0.839 & 0.562 & 0.531 & 7.599 & 6.409 \\
CMA--ES SIMBICON             & 0.154848 & 0.401 & 0.163 & 0.163 & 1.693 & 1.693 \\
PPO Baseline                 & 0.004076 & 0.951 & 0.465 & 0.680 & 7.349 & \textbf{8.869} \\
\bottomrule
\end{tabular}
}
\label{tab:full_performance_comparison}
\end{table*}

We first present an overall quantitative summary of the performance across all tested configurations, before detailing the results for each evaluation scenario in the following
subsections. Table \ref{tab:full_performance_comparison} summarizes the key metrics averaged over all desired speeds, ramp angles, and noise levels. This overview highlights the relative trends among the imitation-decay variants, recurrent architectures, and baselines, providing a unified context for the more detailed analyses that follow. 

As shown in Table \ref{tab:full_performance_comparison}, imitation guided configurations substantially outperform both PPO and the CMA–ES SIMBICON baseline. Introducing imitation decay improves the policy’s balance between accuracy and adaptability: the $\alpha$ = 0.25 model achieves the lowest velocity difference MSE (0.0036), while $\alpha = 0.5$ yields the highest rotation success (0.962). In contrast, the no-decay configuration adapts better in the velocity-difference and rotation demonstrations, maintaining strong stability under mild terrain variations. Disabling reference-state initialization (No RSI)
degrades performance, confirming the importance of phase randomized starts for stable gait learning. 

\begin{figure}[t]
    \centering
    \includegraphics[width=0.75\textwidth]{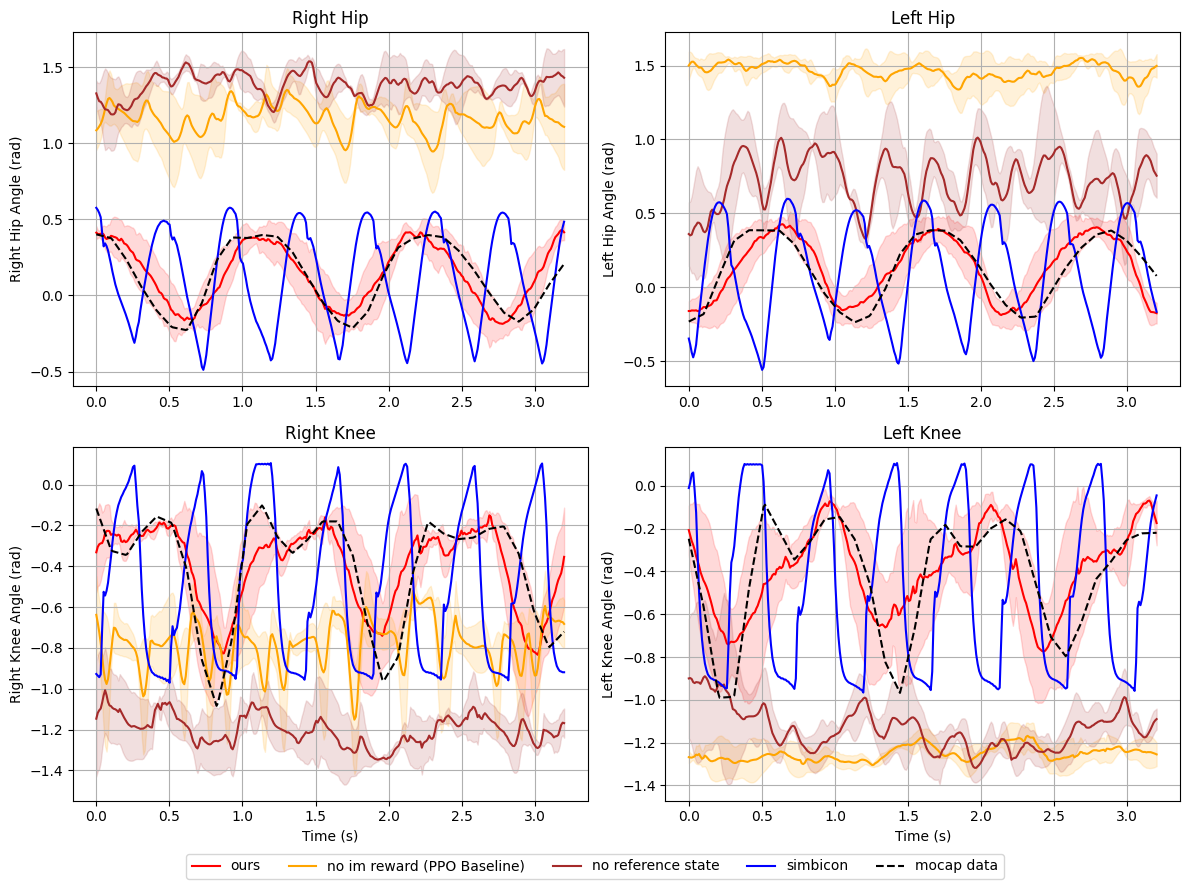}
    \caption{
    Comparison of joint trajectories between our imitation-based controller, ablated variants, and motion-capture reference at $1.2~\text{m/s}$. 
    Each curve represents the mean over three random seeds, and shaded areas indicate $\pm$SD. 
    Motion-capture data correspond to the mean of five trials with speeds between $1.195$ and $1.205~\text{m/s}$.
    }
    \label{fig:jointtraj}
\end{figure}

As illustrated in Fig. \ref{fig:jointtraj}, the joint trajectories generated by our imitation-guided controller closely follow the average motion-capture data across all joints, confirming that the proposed reward formulation effectively preserves human like coordination. In contrast, the no imitation reward configuration (PPO baseline) which retains reference states but omits the imitation reward term and the no reference state configuration—trained without access to reference trajectories both diverge significantly from the human reference. These ablated models yield mechanically unstable and visually unrealistic gaits, lacking the smooth stance swing transitions and phase consistency observed in human motion. Overall, the imitation based policy achieves a precise alignment with the natural kinematic structure of the motion capture gait, demonstrating the necessity of both imitation rewards and reference-state conditioning for stable and human like locomotion.
To test the models performance in rough conditions. We applied 4 different demo scenarios to test the system.

\begin{itemize}
    \item \textbf{Velocity comparison demo}: flat and even surface with varying commanded speeds from $0$ to $2.2\,\mathrm{m/s}$ to compare actual speed vs commanded speed.  
    \item \textbf{Rotation demo}: ramp angles from $-15^\circ$ to $15^\circ$ with commanded speeds from $0$ to $2.2\,\mathrm{m/s}$ to compare the models strength against variable speeds and ramp angles.
    \item \textbf{Noisy plane demo}: stochastic terrains are generated with the method given by \cite{hamzaccebi2024analysis}, with varying $\gamma$ and $\omega$ parameters to construct different noise planes with varying levels of difficulty to test the model's performance.
    \item \textbf{Velocity tracking demo}: commanded speed varied online within a single rollout episode to test transient response and tracking behavior.  
\end{itemize}

Overall results for the each demo for each configuration is presented in table \ref{tab:full_performance_comparison}.
Three different neural network configurations are evaluated, and the first configuration is also tested without RSI and with imitation decay using different $\alpha$ coefficients, in order to observe the effect of the training methodology on the system for ablation purposes. A total of 6 different configurations are tested. SIMBICON control framework, whose parameters are optimized using (CMA-ES) following the procedure of \cite{wang2009optimizing} and PPO baseline are included in the same plots for comparison.

\subsection{Velocity Comparison Demo}
\begin{figure}[t]
\centering
\includegraphics[width=0.7\textwidth]{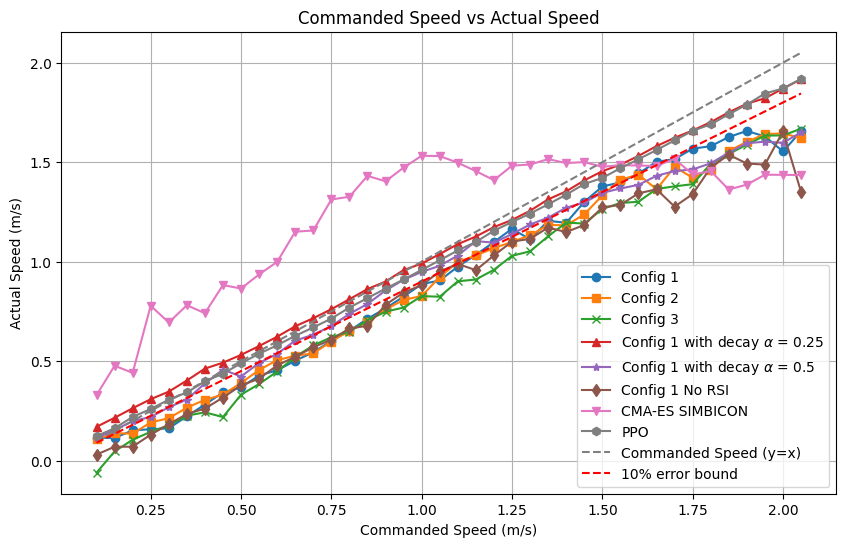}
\caption{Velocity Comparison Demo Results }
\label{veldiff}
\end{figure}

All PPO-based controllers exhibit an approximately linear relationship between commanded and realized speed in the low-to-moderate range, closely following the \(y = x\) reference and remaining within the \(\pm 10\%\) error margin up to about \(1.3\)–\(1.5~\mathrm{m/s}\). Beyond this range, the PPO baseline and imitation-decay configurations maintain the closest alignment with the commanded speed, while the non-decay variant of Configuration~1 still tracks within 10\% error until approximately \(1.8~\mathrm{m/s}\). In contrast, recurrent policies and the No-RSI ablation show degraded performance, with actual velocities deviating substantially from their targets. The CMA--ES SIMBICON baseline fails to capture the desired trend altogether. Overall, the scheduled imitation-decay mechanism enhances high-speed tracking and linearity, whereas proper regularization and randomized state initialization help prevent early saturation and preserve generalization. Results can be seen in Fig. \ref{veldiff}

\subsection{Rotation Demo}

\begin{figure}[t]
\centering
\includegraphics[width=0.7\textwidth]{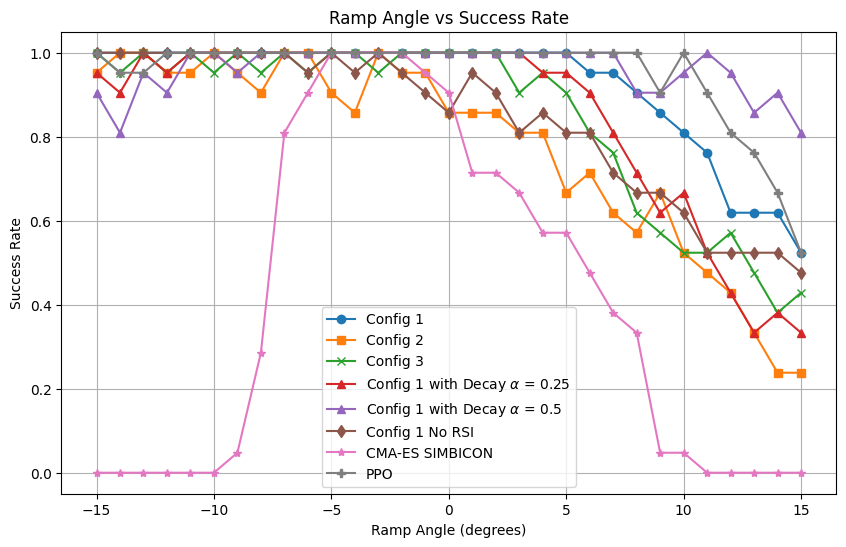}
\caption{Average success rate across ramp angles}
\label{rotdemo}
\end{figure}

Performance across inclined terrain, varying the ramp angle from $-15^\circ$ to $15^\circ$ while sweeping commanded speeds up to $2.2~\mathrm{m/s}$ are evaluated.  
Fig.~\ref{rotdemo} reports the average success rate across angles.

Success remains nearly perfect on declines: for $-15^{\circ} \le \theta \le 0^{\circ}$, all PPO-based controllers maintain success rates close to unity with only minor variability. Notably, although training exposures were limited to ramp angles within $[-5^{\circ}, +5^{\circ}]$, the learned policies generalize effectively to steeper inclinations, preserving high success beyond the training range. The PPO baseline, Configuration~1, and its imitation-decay variants sustain high success rates up to $5^{\circ}$ and remain the most stable beyond that threshold. Among them, the imitation-decay configuration with $\alpha = 0.5$ achieves the highest success rate, retaining approximately $80\%$ success even at $15^{\circ}$. In contrast, recurrent policies and the No-RSI ablation exhibit degraded performance under steep inclines. The CMA--ES SIMBICON baseline demonstrates stability only near flat terrain and mild declines, failing consistently beyond $5^{\circ}$. Overall, these results indicate that PPO-based variants exhibit superior robustness to unseen slope conditions, while imitation decay enhances stability and the RSI mechanism improves overall generalization.

\subsection{Noisy Plane Demo}

\begin{figure}[t]
\centering
\begin{minipage}{0.45\linewidth}
    \centering
    \includegraphics[width=\linewidth]{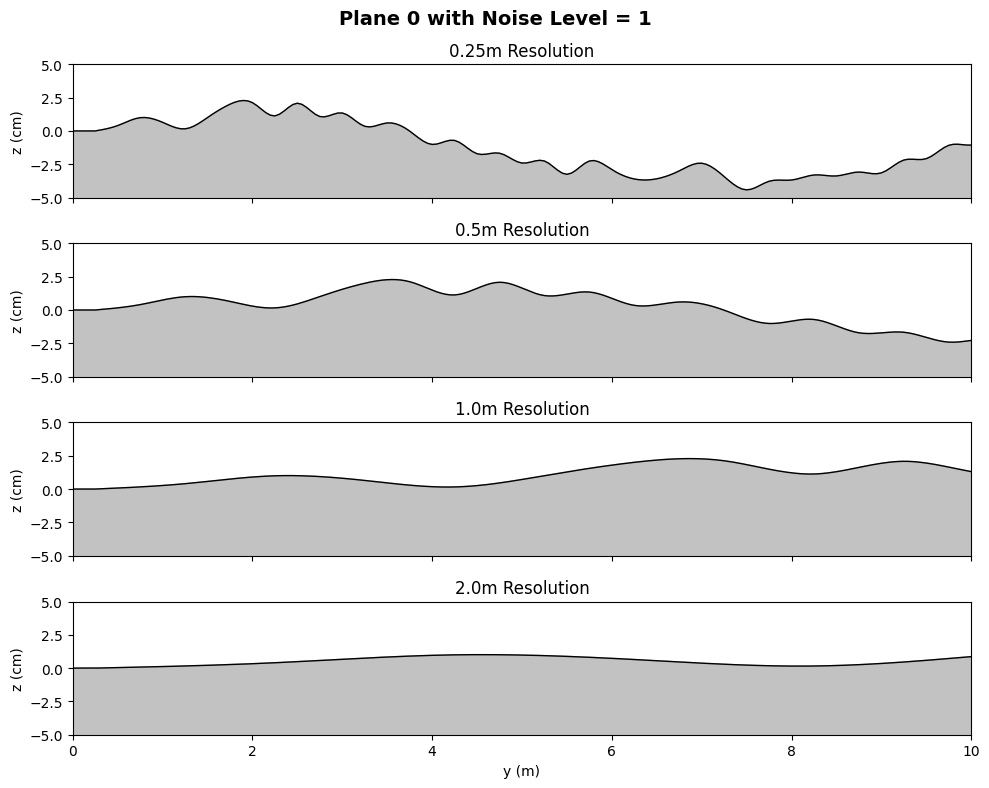}
\end{minipage}
\begin{minipage}{0.45\linewidth}
    \centering
    \includegraphics[width=\linewidth]{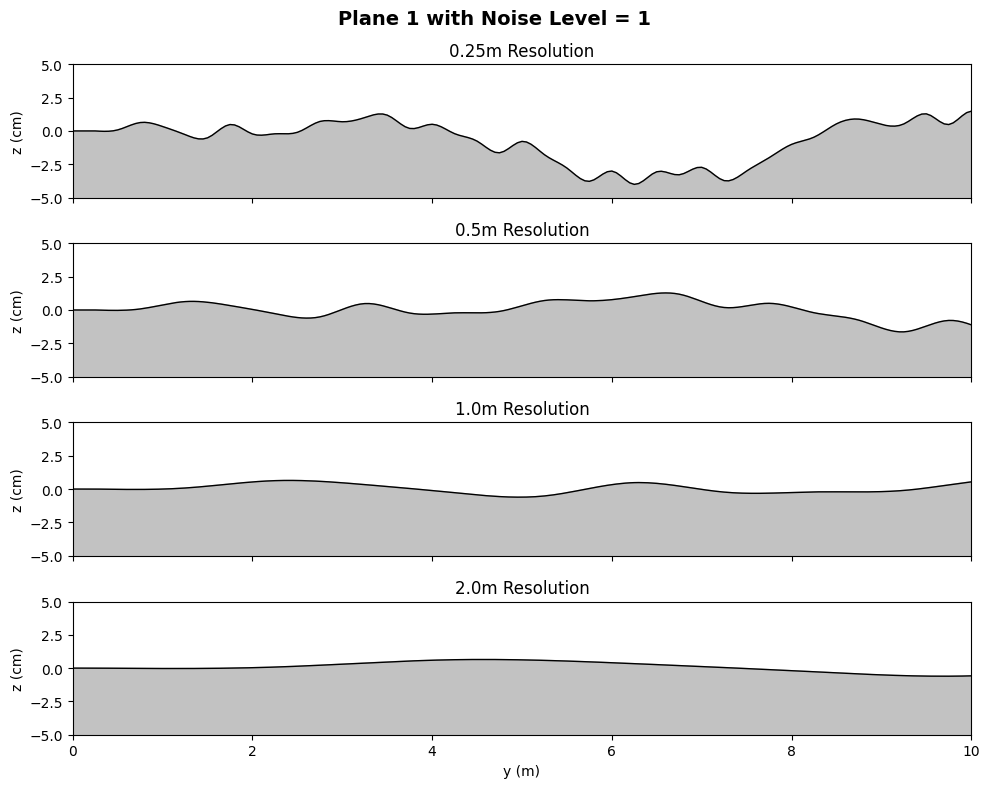}
\end{minipage}
\caption{Noisy plane profiles for Plane~0 and Plane~1.}
\label{fig:noisy_planes_side}
\end{figure}

Two noisy planes are created randomly using the same function in \cite{hamzaccebi2024analysis}. Each noise plane profile is adjusted by changing the $\omega$ (Noise Level) the number used to multiply the height level of the plane and the $\gamma$ (Noise Resolution) the resolution of each noise level change. $\omega$ is changed from 1 to 20 to increase the roughness of the plane while $\gamma$ is changed to $[0.25,0.5,1.0,2.0]$ meters to increase the smoothness of the planes. Effects of increasing $\gamma$ can be seen in Fig. \ref{fig:noisy_planes_side}.

\begin{figure}[htbp]
  \centering
    \includegraphics[width=0.7\textwidth]{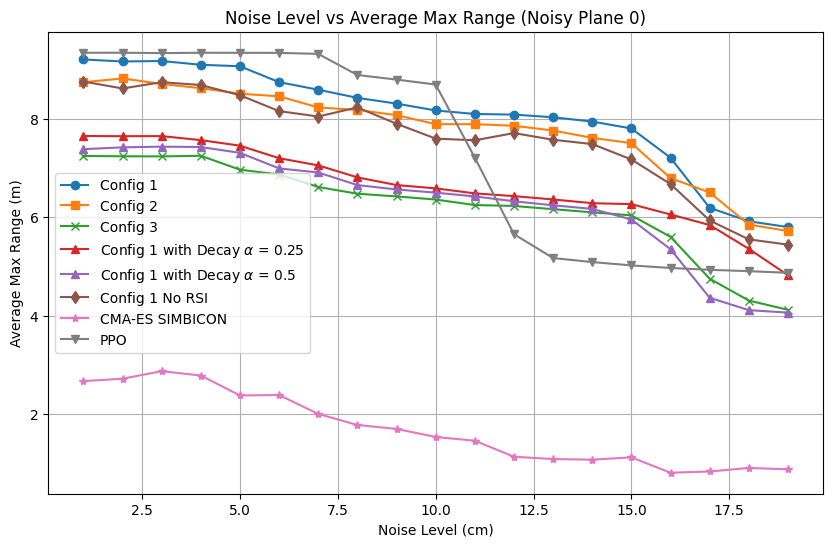}
    \caption{Noisy Plane~0: average max range vs.~noise amplitude.}
    \label{fig:noisy0-range}
\end{figure}

\begin{figure}[htbp]
  \centering
    \includegraphics[width=0.7\textwidth]{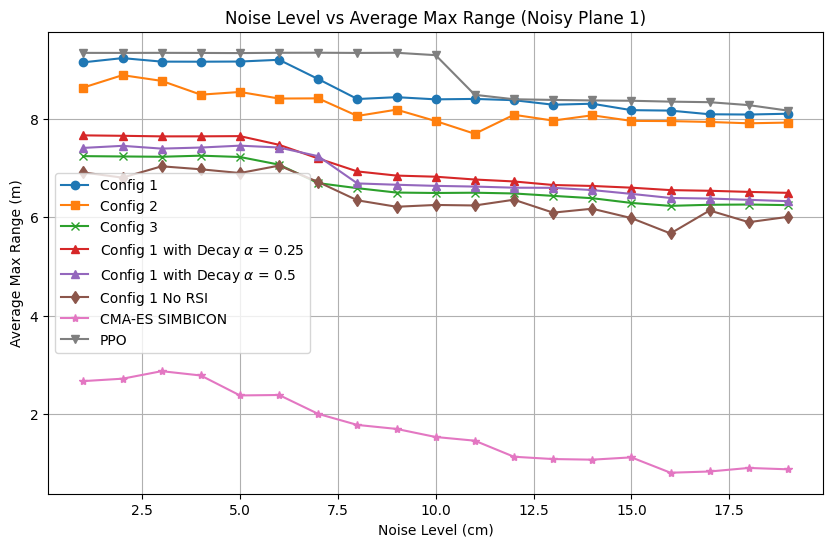}
    \caption{Noisy Plane~1: average max range vs.~noise amplitude.}
    \label{fig:noisy1-range}
\end{figure}

Figures~\ref{fig:noisy0-range} and~\ref{fig:noisy1-range} illustrate the average maximum travel range before failure as a function of terrain noise amplitude for two independent noisy-plane realizations. All policies were trained exclusively on flat ground; during evaluation, the terrain height was perturbed using spatially correlated noise with amplitudes ranging from \(1\) to \(20~\mathrm{cm}\). Across both planes, Configuration~1 consistently achieves the largest traversal ranges and exhibits the most graceful degradation as surface roughness increases. Recurrent policies remain competitive at low amplitudes $(\lesssim 8~\mathrm{cm})$ but underperform at higher noise levels, while both imitation-decay and No-RSI ablations show reduced robustness. The PPO baseline performs well at lower noise levels but experiences a sharp performance drop beyond \(10~\mathrm{cm}\) for Noisy Plane~0. In contrast, the CMA--ES SIMBICON baseline collapses early, maintaining ranges of only \(\approx 2\!-\!3~\mathrm{m}\) even under mild perturbations, confirming its limited tolerance to high-frequency terrain variations. Differences in plane sensitivity are evident: Noisy Plane~0 imposes harsher local gradients, whereas Noisy Plane~1 yields smoother profiles and higher sustained ranges. Overall, PPO-based controllers generalize effectively to rough surfaces without explicit noise-domain training, maintaining substantially greater operational ranges than both the PPO baseline and the SIMBICON reference. The weaker performance of the imitation-decay variants further suggests that direct human imitation enhances stability, while the abrupt degradation of PPO on Noisy Plane~0 supports this interpretation.

\subsection{Velocity Tracking Demo}

\begin{figure}[htbp]
\centering
\includegraphics[width=0.7\textwidth]{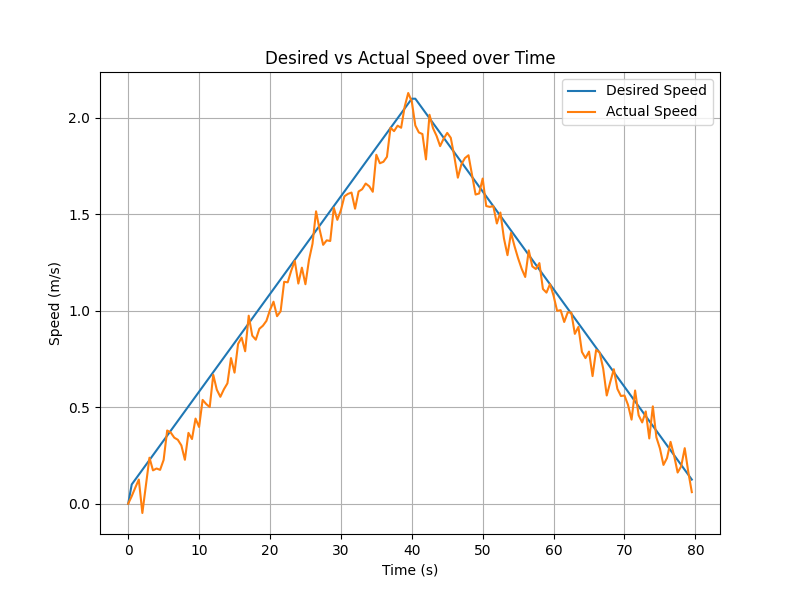}
\caption{Velocity Tracking Demo: desired vs actual velocity during a time-varying speed profile}
\label{fig:trackdemo}
\end{figure}

To test the dynamic adaptability, we varied the commanded speed within a single rollout episode (ramp-up to $2.0\,\mathrm{m/s}$, then back down). Figure \ref{fig:trackdemo} shows that Configuration 1 smoothly tracks the changing command with small lag ($\sim0.1$--$0.2\,\mathrm{m/s}$).

%===============================================================================
\section{Conclusion}

This work has presented a two-stage locomotion framework for a 7-DoF planar biped: a frequency-domain gait generator, trained on motion-capture data with morphology and speed conditioning, followed by a torque-controlled PPO policy that refines these compact gait references via a blended imitation and task reward. By combining the structured periodicity from gait generator network with on-policy RL, we achieve walking behaviour that is both robust across speeds, slopes and terrain irregularities, and natural in gait appearance. Our experiments show that the proposed controller outperforms the SIMBICON and PPO baselines in terms of velocity range, slope tolerance and disturbance robustness.

Our main contributions are as follows: we first introduce a novel frequency-domain gait generator network that maps morphology and desired speed to full joint trajectories; we secondly integrate this gait generator into a hybrid RL framework, where the PPO controller uses the generated trajectories as references and learns to refine them under task and energy constraints; and thirdly we propose a training methodology combining RSI and an imitation-decay schedule, which together accelerate training, prevent reward-hacking and improve generalization on unseen terrain. 

To the best of our knowledge, no prior work has trained a neural gait-generator in the frequency (FFT) domain from motion-capture data and used it as the reference generator for RL-based bipedal gait learning. This approach therefore offers a niche yet straightforward and implementable solution that bridges imitation priors and reinforcement learning in a compact and general-purpose gait-generation module. 

In future work we will extend the framework to full 3D humanoid models, equip it with richer sensory feedback and contact-dynamics.

%===============================================================================
\newpage
\bibliographystyle{plainnat}
\bibliography{ifacconf}

\begin{thebibliography}{19}
\providecommand{\natexlab}[1]{#1}
\providecommand{\url}[1]{\texttt{#1}}
\expandafter\ifx\csname urlstyle\endcsname\relax
  \providecommand{\doi}[1]{doi: #1}\else
  \providecommand{\doi}{doi: \begingroup \urlstyle{rm}\Url}\fi

\bibitem[Bao et~al.(2024)Bao, Humphreys, Peng, and Zhou]{surveypaper}
Lingfan Bao, Joseph Humphreys, Tianhu Peng, and Chengxu Zhou.
\newblock Deep reinforcement learning for bipedal locomotion: A brief survey.
\newblock \emph{arXiv preprint arXiv:2404.17070}, 2024.

\bibitem[{\c{C}}atalba{\c{s}} and Morg{\"u}l(2022)]{ccatalbacs2022two}
Bahad{\i}r {\c{C}}atalba{\c{s}} and {\"O}mer Morg{\"u}l.
\newblock Two-legged robot motion control with recurrent neural networks.
\newblock \emph{Journal of Intelligent \& Robotic Systems}, 104\penalty0 (4):\penalty0 59, 2022.

\bibitem[Hamza{\c{c}}ebi et~al.(2024)Hamza{\c{c}}ebi, Uyanik, and Morg{\"u}l]{hamzaccebi2024analysis}
Hasan Hamza{\c{c}}ebi, Ismail Uyanik, and {\"O}mer Morg{\"u}l.
\newblock On the analysis and control of a bipedal legged locomotion model via partial feedback linearization.
\newblock \emph{Bioinspiration \& Biomimetics}, 19\penalty0 (5):\penalty0 056004, 2024.

\bibitem[Itahashi et~al.(2024)Itahashi, Itoh, Fukumoto, and Wakuya]{itahashi2024reinforcement}
Naoya Itahashi, Hideaki Itoh, Hisao Fukumoto, and Hiroshi Wakuya.
\newblock Reinforcement learning of bipedal walking using a simple reference motion.
\newblock \emph{Applied Sciences}, 14\penalty0 (5):\penalty0 1803, 2024.

\bibitem[Krishna et~al.(2021)Krishna, Mishra, Castillo, Hereid, and Kolathaya]{krishna2021learning}
Lokesh Krishna, Utkarsh~A Mishra, Guillermo~A Castillo, Ayonga Hereid, and Shishir Kolathaya.
\newblock Learning linear policies for robust bipedal locomotion on terrains with varying slopes.
\newblock In \emph{2021 IEEE/RSJ International Conference on Intelligent Robots and Systems (IROS)}, pages 5159--5164. IEEE, 2021.

\bibitem[Li et~al.(2025)Li, Peng, Abbeel, Levine, Berseth, and Sreenath]{li2025reinforcement}
Zhongyu Li, Xue~Bin Peng, Pieter Abbeel, Sergey Levine, Glen Berseth, and Koushil Sreenath.
\newblock Reinforcement learning for versatile, dynamic, and robust bipedal locomotion control.
\newblock \emph{The International Journal of Robotics Research}, 44\penalty0 (5):\penalty0 840--888, 2025.

\bibitem[Peng et~al.(2025)Peng, Bao, and Zhou]{peng2025gait}
Tianhu Peng, Lingfan Bao, and CHengxu Zhou.
\newblock Gait-conditioned reinforcement learning with multi-phase curriculum for humanoid locomotion.
\newblock \emph{arXiv preprint arXiv:2505.20619}, 2025.

\bibitem[Peng et~al.(2018)Peng, Abbeel, Levine, and Van~de Panne]{peng2018deepmimic}
Xue~Bin Peng, Pieter Abbeel, Sergey Levine, and Michiel Van~de Panne.
\newblock Deepmimic: Example-guided deep reinforcement learning of physics-based character skills.
\newblock \emph{ACM Transactions On Graphics (TOG)}, 37\penalty0 (4):\penalty0 1--14, 2018.

\bibitem[Peng et~al.(2020)Peng, Coumans, Zhang, Lee, Tan, and Levine]{peng2020learning}
Xue~Bin Peng, Erwin Coumans, Tingnan Zhang, Tsang-Wei Lee, Jie Tan, and Sergey Levine.
\newblock Learning agile robotic locomotion skills by imitating animals.
\newblock \emph{arXiv preprint arXiv:2004.00784}, 2020.

\bibitem[Schreiber and Moissenet(2019)]{schreiber2019multimodal}
C{\'e}line Schreiber and Florent Moissenet.
\newblock A multimodal dataset of human gait at different walking speeds established on injury-free adult participants.
\newblock \emph{Scientific data}, 6\penalty0 (1):\penalty0 111, 2019.

\bibitem[Schulman et~al.(2015)Schulman, Moritz, Levine, Jordan, and Abbeel]{gae}
John Schulman, Philipp Moritz, Sergey Levine, Michael Jordan, and Pieter Abbeel.
\newblock High-dimensional continuous control using generalized advantage estimation.
\newblock \emph{arXiv preprint arXiv:1506.02438}, 2015.

\bibitem[Schulman et~al.(2017)Schulman, Wolski, Dhariwal, Radford, and Klimov]{schulman2017proximal}
John Schulman, Filip Wolski, Prafulla Dhariwal, Alec Radford, and Oleg Klimov.
\newblock Proximal policy optimization algorithms.
\newblock \emph{arXiv preprint arXiv:1707.06347}, 2017.

\bibitem[Su and Huang(2023)]{su2023multi}
Mengya Su and Yan Huang.
\newblock Multi-speed walking gait generation for bipedal robots based on reinforcement learning and human motion imitation.
\newblock In \emph{2023 42nd Chinese Control Conference (CCC)}, pages 4815--4821. IEEE, 2023.

\bibitem[Taga et~al.(1991)Taga, Yamaguchi, and Shimizu]{taga1991self}
Gentaro Taga, Yoko Yamaguchi, and Hiroshi Shimizu.
\newblock Self-organized control of bipedal locomotion by neural oscillators in unpredictable environment.
\newblock \emph{Biological cybernetics}, 65\penalty0 (3):\penalty0 147--159, 1991.

\bibitem[Wang et~al.(2009)Wang, Fleet, and Hertzmann]{wang2009optimizing}
Jack~M Wang, David~J Fleet, and Aaron Hertzmann.
\newblock Optimizing walking controllers.
\newblock In \emph{ACM SIGGRAPH Asia 2009 papers}, pages 1--8. 2009.

\bibitem[Wensing and Orin(2013)]{slip}
Patrick~M Wensing and David~E Orin.
\newblock High-speed humanoid running through control with a 3d-slip model.
\newblock In \emph{2013 IEEE/RSJ International Conference on Intelligent Robots and Systems}, pages 5134--5140. IEEE, 2013.

\bibitem[Xie et~al.(2018)Xie, Berseth, Clary, Hurst, and Van~de Panne]{xie2018feedback}
Zhaoming Xie, Glen Berseth, Patrick Clary, Jonathan Hurst, and Michiel Van~de Panne.
\newblock Feedback control for cassie with deep reinforcement learning.
\newblock In \emph{2018 IEEE/RSJ International Conference on Intelligent Robots and Systems (IROS)}, pages 1241--1246. IEEE, 2018.

\bibitem[Yin et~al.(2007)Yin, Loken, and Van~de Panne]{yin2007simbicon}
KangKang Yin, Kevin Loken, and Michiel Van~de Panne.
\newblock Simbicon: Simple biped locomotion control.
\newblock \emph{ACM Transactions on Graphics (TOG)}, 26\penalty0 (3):\penalty0 105--es, 2007.

\bibitem[Zhang et~al.(2025)Zhang, Zhang, Chen, Chen, Wang, and Xiong]{zhang2025natural}
Haodong Zhang, Liang Zhang, Zhenghan Chen, Lu~Chen, Yue Wang, and Rong Xiong.
\newblock Natural humanoid robot locomotion with generative motion prior.
\newblock \emph{arXiv preprint arXiv:2503.09015}, 2025.

\end{thebibliography}

\end{document}